\documentclass[twoside,11pt]{article}

\usepackage{amsthm}

\usepackage{jmlr2e}
\usepackage[margin=1in]{geometry}

\usepackage{wrapfig}

\usepackage{subcaption}
\usepackage{xcolor}
\usepackage{microtype}
\usepackage[font=small,labelfont=bf]{caption}
\usepackage{booktabs}
\usepackage{lipsum}
\usepackage{amsmath}
\usepackage{xspace}
\usepackage{enumitem}
\usepackage{hyperref}
\usepackage{natbib}

\firstpageno{1}

\newcommand{\westerngroup}{\ensuremath{G_0}\xspace}
\newcommand{\nonwesterngroup}{\ensuremath{G_1}\xspace}

\newcommand{\preperiod}{\ensuremath{t_{\textrm{SB}}}\xspace}
\newcommand{\postperiod}{\ensuremath{t_{\textrm{DB}}}\xspace}

\newcommand{\rating}{\ensuremath{r_{ij}}\xspace}

\newcommand{\paperyear}{\ensuremath{T_j}\xspace}
\newcommand{\papergroup}{\ensuremath{S_j}\xspace}

\newcommand{\disparity}[1]{\ensuremath{\Delta_{#1}}\xspace}
\newcommand{\doubleblinddisparity}[1]{\ensuremath{\Delta_{#1}^{\textrm{DB}}}\xspace}
\newcommand{\singleblinddisparity}[1]{\ensuremath{\Delta_{#1}^{\textrm{SB}}}\xspace}
\newcommand{\bias}[1]{\ensuremath{\doubleblinddisparity{#1} - \singleblinddisparity{#1}}\xspace}
\newcommand{\perf}[1]{\ensuremath{\textrm{perf}(#1)}\xspace}
\newcommand{\ratingsbias}{\ensuremath{\gamma}\xspace}
\newcommand{\ratingsdisparity}[1]{\ensuremath{\Delta_{#1}^\textrm{rating}}\xspace}

\newtheorem{assumption}{Assumption}

\begin{document}

\title{Uncovering Latent Biases in Text:\\ Method and Application to Peer Review\vspace{-27.5mm}}

\author{}

\editor{}

\maketitle

\begin{centering}
  Emaad Manzoor$^\ast$, Nihar B. Shah$^\dagger$\\
  Carnegie Mellon University\\
  \texttt{$^\ast$emaad@cmu.edu, $^\dagger$nihars@cs.cmu.edu}\\
\end{centering}

\vspace{8.5mm}

\begin{abstract}%
  Quantifying systematic disparities in numerical quantities such as employment rates and wages between population subgroups
  provides compelling evidence for the existence of societal biases.
  However, biases in the \emph{text} written for members of different subgroups (such as in recommendation letters for male and non-male candidates), though widely reported anecdotally, remain challenging to quantify.
  In this work, we introduce a novel framework to quantify bias in text
  caused by the visibility of subgroup membership indicators.
  We develop a nonparametric estimation and inference procedure to estimate this bias. We then formalize an identification strategy to causally link the estimated bias to the visibility of subgroup membership indicators, provided observations from time periods both before and after an identity-hiding policy change.
  We identify an application wherein ``ground truth'' bias can be inferred to evaluate our framework, instead of relying on synthetic or secondary data.
  Specifically, we
  apply our framework to quantify biases in the text of peer reviews from a reputed machine learning conference before and after the conference adopted a double-blind reviewing policy.
  We show evidence of biases in the review ratings that serves as ``ground truth'', and show that our proposed framework accurately detects these biases from the review text \textit{without} having access to the review ratings.
\end{abstract}

\section{Introduction}
\label{sec:intro}

Societal biases against individuals based on race, gender, and other attributes can lead to disparities in hiring \citep{bertrand2004emily}, wages \citep{blau2017gender} and incarceration rates \citep{alesina2014test}, among other socioeconomic outcomes. Uncovering evidence of such biases informs the creation of policies to eliminate long-standing gaps between different population subgroups.

Many important socioeconomic outcomes are influenced by written feedback, such as academic hiring that is influenced by reference letters, and employee appraisals that rely on  managerial performance reviews. Previous studies have reported biases in the \emph{text} of such feedback \citep{mitchell2018gender,madera2019raising,altomonte2020exploiting}.
\citet{mitchell2018gender}
find that student evaluations of female faculty teaching were more likely to focus on communication ability. \citet{madera2019raising} find that academic letters of recommendation for women were more likely to contain ``doubt-raising'' phrases. In the context of managerial feedback provided during employee appraisals, \citet{altomonte2020exploiting} find that ``women were more likely to receive vague feedback that did not offer specific details of what they had done well and what they could do to advance''.

Such biases in text can have severe economic consequences. For example, differences in the media framing of natural disasters were found to be associated with large disparities in the amount of allocated foreign aid \citep{stromberg2007natural,guardian2014}. Similarly, online reviews of Asian and Indian restaurants declared them ``unauthentic'' when they defied the negative stereotypes of uncleanliness, leading to significant losses in revenue \citep{eater2019}.

Biases in text are more prevalent than those in numerical feedback due to the absence of enforced structure \citep{hbr2019}, and are more difficult to detect due to their subtle manner of expression \citep{morstatter2018identifying}. In addition, the inherently unstructured nature of text makes it challenging to quantify the biases it contains. Without the ability to quantify and provide credible evidence of such biases, society remains at risk of widening socioeconomic gaps fueled by the unchecked expression of prejudices in written feedback.

In this work, we propose a framework to quantify biases in text, provided data from time periods both before and after an identity-hiding policy change. Our proposed framework extends the difference-in-differences causal inference methodology introduced by \citet{card1994minimum} (which compares differences in \emph{numerical} quantities over time) to handle differences in \emph{unstructured text} over time. We motivate and evaluate our framework in the setting of scholarly peer review, which is a key mechanism of feedback and quality assurance in scientific research. Specifically, we assemble a dataset of peer reviews from the International Conference of Learning Representations (ICLR), which switched from single-blind to double-blind reviewing in 2018. We test for biases in the peer review text, which is prone to prejudice due to the lack of enforced structure. Importantly, our dataset enables estimating ``ground truth'' biases using the peer review ratings (a numerical quantity). The goal of our proposed framework is to quantify biases in the peer review text that are consistent with the ``ground truth'', without having access to the review ratings.

\noindent Our main contributions are:
\begin{enumerate}[label=\Roman*., topsep=12pt, itemsep=12pt]
  \item We formalize bias as a \textit{causal estimand} --- the disparity in the peer review text between the subgroups \textit{caused} by the visibility of author identities --- that relies on a weaker assumption than ``no unobserved confounders''. We propose a nonparametric estimation and inference procedure to quantify this bias. Our procedure makes no assumptions on the data-generating process of the peer review text and requires no feature engineering.

  \item We apply our proposed framework to quantify the bias in the text of peer reviews from the International Conference on Learning Representations (ICLR). We detect a statistically significant bias with respect to the authors' \emph{affiliation}, but find no evidence of bias with respect to the authors' \emph{perceived gender}.

  \item
  Our chosen application is motivated by the opportunity to evaluate our proposed framework on ``ground truth'' biases derived from review ratings. Specifically, we evaluate our proposed framework by comparing the estimated biases in the peer review text with the biases in the peer review ratings estimated using the difference-in-differences methodology \citep{card1994minimum,angrist2008mostly}. We show
  that the biases in the peer review text estimated using our proposed framework are consistent with the ``ground truth''.
\end{enumerate}

In Appendix \ref{sec:alternativedisparity}, we also evaluate an alternative measure of disparity in the text proposed by \citet{gentzkow2019measuring}, and this empirical evaluation reveals that  our proposed measure of disparity has greater statistical power.
Finally, although presented in the context of peer review, our proposed framework can also be applied to other settings such as testing for biases in feedback provided by employers when reviewing employees \citep{goldin2000orchestrating} and evaluating potential hires \citep{capowski1994ageism}. 

\noindent Replication code and data for this work is available at 
\url{https://emaadmanzoor.com/biases-in-text/}.

\section{Related Work}
There is recently an increasing focus on designing methods for improving the fairness and quality of peer review, and our work contributes to this important line of literature~\citep{Garg2010papers,roos2011calibrate,cabanac2013capitalizing,lee2015commensuration,xu2018strategyproof,kang2018dataset,wang2018your,stelmakh2018forall,noothigattu2018choosing,kobren19localfairness,fiez2020super,stelmakh2020catch,jecmen2020manipulation}. 
Our work complements previous studies that investigated biases in peer review~\citep{goldberg1968women,ceci1982peer,swim1989joan,lloyd1990gender,blank1991effects,garfunkel1994effect,snodgrass2006single,ross2006effect,budden08dbfemale,webb2008does,walker2015bias,okike2016single,seeber2017does,bernard2018gender,tomkins2017reviewer,stelmakh2019testing,salimi2020causal}. In contrast with our work, the aforementioned studies focused on biases in numerical quantities such as acceptance rates and review ratings caused by visible author identities.

Our approach compares single-blind and double-blind reviews for gender and affiliation-based population subgroups.
A number of previous studies have quantified biases using the peer-review ratings provided for different subgroups under single-blind and double-blind reviewing policies.
\citet{blank91effects} conducts a randomized control trial and finds no significant difference in the review ratings for male and female authors under single and double-blind reviewing.
~\citet{ross2006effect} compare single and double-blind reviewing in different years at a medical conference and find that the association between abstract acceptance and whether the authors were affiliated to institutions in the USA reduces significantly when reviewing is double-blind.
\citet{madden2006impact} compare single and double-blind reviewing for the SIGMOD conference in different years and find that the mean number of accepted of papers by a ``prolific'' author remained largely similar before and after the switch to double-blind reviewing. In contrast, \citet{tung2006impact} find a significant reduction in the \emph{median} number of accepted papers by a ``prolific'' author after SIGMOD switched to double-blind reviewing.
More recently, \citet{tomkins2017reviewer} conduct a semi-randomized controlled trial with the WSDM 2017 conference and do not find a significant association between a paper's single-blind review rating and whether its authors were women, or whether its authors were affiliated to institutions in the USA. \citet{salimi2020causal} compare single and double-blind reviewing in several conferences and find a significant effect of institutional prestige on review ratings when reviewing was single-blind (and none when reviewing was double-blind). In the context of single-blind journal peer review, \cite{thelwall2020does} find that reviewers are biased in favor of authors from their home country. Our work complements these studies by focusing on biases in the review \emph{text}, instead of on biases in numerical quantities.

Our proposed framework is closely related to the bias discovery method proposed by \citet{field2020unsupervised}, who train a machine learning model to identify differences in online comments addressed towards men and women. Their method also relies on the idea that text which is predictive of gender is likely to contain bias. They show that their method can detect gender bias on a labeled dataset from a different domain than the one their model was trained on.
However, the method proposed by \citet{field2020unsupervised} crucially depends on the ``no unobserved confounders'' assumption.
This assumption is restrictive and unlikely to hold in practice. In contrast, our proposed framework relies on a weaker assumption that remains valid under a large class of unobserved confounders (though requiring additional data, from two different time periods).
We also overcome a key limitation in the \citep{field2020unsupervised} by evaluating our proposed framework on ``ground truth'' derived from the same peer review process used for bias estimation, instead of on secondary datasets or tasks.

\section{Problem Definition}
\label{sec:problem}

We assume the availability of peer reviews from a conference in two different years, with author identities (their names, emails, and affiliations) visible to reviewers during the peer review process in exactly one of the two years. Our goal is to quantify biases in the \textit{text}
of the peer reviews written for papers belonging to two pre-specified subgroups based on a selected identifying attribute of their authors.

Previous studies have reported systematic \textit{disparities} in the text written for different population subgroups. For example, \citet{madera2019raising} report that recommendation letters for women are more likely to contain ``doubt-raising'' language then those for men. However, such disparities by themselves are not  sufficient evidence of bias \citep{rathore2004differences}. While disparities are observed differences in the review text, biases are observed differences that are \textit{caused} by author identity visibility, and not other factors.

In a \textit{counterfactual} universe where the author identities are hidden, the bias must be zero (since its cause no longer exists) but the observed disparity can be nonzero.
For example, if we partition papers into subgroups based on their first author's affiliation country, disparities in the review text could also arise due to country-specific preferences for different research topics.
When defining bias, our goal is to separate the disparity caused by author identity visibility from the disparity caused by other factors. We now formalize bias as a causal estimand with the potential outcomes framework \citep{imbens2015causal}, and derive an expression that relates bias to the disparities observed in different time periods.

Consider papers submitted to a conference in the years \preperiod and \postperiod, where the conference employed single-blind reviewing in year \preperiod and double-blind reviewing in year \postperiod. We partition the papers into two subgroups $\westerngroup$ and $\nonwesterngroup$
based on a selected identifying attribute of their authors (such their affiliation or perceived gender). Our goal is to formalize bias as the disparity in the review text for papers in each subgroup caused by the visibility of this identifying attribute to reviewers during the peer review process.

We denote by \disparity{t} the \textit{observed disparity} in the review text in year $t \in \{\preperiod, \postperiod\}$. While we propose a careful nonparametric formulation of \disparity{t} in Section~\ref{sec:proposed},
\disparity{t} can be viewed as any measure of the difference in the review text in year $t$ between subgroups $\westerngroup$ and $\nonwesterngroup$.
We denote by \singleblinddisparity{t} and \doubleblinddisparity{t}
the \textit{counterfactual disparities} in the review text in year $t$ that would have been observed had author identities been visible to and hidden from reviewers, respectively. Only one of the quantities \doubleblinddisparity{t} and \singleblinddisparity{t} is visible in each year. When $t = \preperiod$ and reviewing was single-blind, \doubleblinddisparity{\preperiod} is unobserved and quantifies the disparity in year \preperiod had reviewing been double-blind instead.
When $t = \postperiod$ and reviewing was double-blind, \singleblinddisparity{\postperiod} is unobserved and quantifies the disparity in year $\postperiod$ had reviewing been single-blind instead.

We define the \textit{bias} (our causal estimand) as a difference in counterfactual disparities:
\begin{align}
  \textrm{bias} = \bias{\postperiod}.
  \label{eq:bias}
\end{align}
Eq.~\eqref{eq:bias} subtracts the disparity \singleblinddisparity{\postperiod} (caused by both author identity visibility and other factors) from the disparity \doubleblinddisparity{\postperiod} (caused by other factors only) to isolate the disparity caused by author identity visibility only.
Note that the bias could also have been defined as \bias{\preperiod} (the change in the disparity that would have been observed had reviewing in year \preperiod been double-blind instead). Either definition is valid and applicable to our framework (after minor algebraic changes).

Note that our definition of bias does not restrict how the subgroups are defined. As such, the bias with respect to subgroups based on different identifying attributes that lead to the same partitions \westerngroup and \nonwesterngroup will be identical, and must be interpreted based on domain expertise. For example, the bias with respect to subgroups based on the authors' seniority would be similar to the bias with respect to subgroups based on the authors' ``fame'' if  senior authors are likely to have had more time to publish, engage with their community and develop connections than junior authors. In practice, we recommend defining subgroups based on hypotheses grounded in socioeconomic theory, and disentangling the biases with respect to different subgroup definitions using institutional knowledge. 

\section{Proposed Framework}
\label{sec:proposed}

Our goal is to estimate the bias defined in Eq.~\eqref{eq:bias} given the peer reviews from a conference in years \preperiod and \postperiod, with author identities visible to reviewers during the peer review process in year \preperiod and hidden in year \postperiod. In this section, we first provide an identification proof to link the causal estimand in Eq.~\eqref{eq:bias} (that contains unobservable counterfactual quantities) with an empirical estimand (that contains observable quantities only). We then propose a nonparametric estimation and inference procedure to estimate the bias from the available peer review data.

\subsection{Identification}
\label{sec:identification}

The bias as defined in Eq.~\eqref{eq:bias} contains the unobservable counterfactual disparity \singleblinddisparity{\postperiod} (the disparity in year \postperiod had reviewing been single-blind), and cannot be estimated without further assumptions; this is the fundamental problem of causal inference \citep{holland1986statistics}. The process of linking a causal estimand defined in terms of unobserved counterfactual quantities with an empirical estimand defined in terms of observed quantities is called \textit{identification}, and relies on one or more \textit{identification assumptions}. We make the following identification assumption:

\begin{assumption}
  The disparity in $t = \postperiod$ had author identities been visible is equal to the disparity in $t = \preperiod$ when author identities were indeed visible: $\singleblinddisparity{\postperiod} = \singleblinddisparity{\preperiod}$.
  \label{assumption:parallel-trends}
\end{assumption}
Assumption~\ref{assumption:parallel-trends} implies that the change in disparity from year \preperiod to \postperiod was caused only by the author identities being hidden in year \postperiod, and not other factors.
Assumption~\ref{assumption:parallel-trends} remains valid in the presence unobserved confounders that affect the review text and (i) that do not vary from \preperiod to \postperiod, or (ii) that vary from \preperiod to \postperiod but affect the review text for both subgroups identically (such as a more critical reviewer pool in year \postperiod). Hence, it is less restrictive than the
``no unobserved confounders'' assumption in prior work \citep{field2020unsupervised}. We further discuss the validity of Assumption~\ref{assumption:parallel-trends} for our setting in Appendix~\ref{sec:validity}.

Let \disparity{t} be the \emph{observed} disparity in year $t \in \{\preperiod,\postperiod \}$. Given Assumption~\ref{assumption:parallel-trends}, we link the bias (that contains an unobservable counterfactual disparity) with an empirical estimand (that contains observable disparities only) with the following \textit{identification proof}:
\begin{align}
 \textrm{bias}
 \stackrel{(i)}{=} \doubleblinddisparity{\postperiod} - \singleblinddisparity{\postperiod}
 \stackrel{(ii)}{=}
  \doubleblinddisparity{\postperiod} - \singleblinddisparity{\preperiod}
  \stackrel{(iii)}{=}
\disparity{\postperiod} - \disparity{\preperiod}
\label{eq:identification}
\end{align}
where the equation (i) follows from the definition in Eq.~\eqref{eq:bias}, equation (ii) follows from Assumption~\ref{assumption:parallel-trends}, and equation (iii) follows from the fact that reviewing was indeed double-blind in year \postperiod ($\doubleblinddisparity{\postperiod}=\disparity{\postperiod}$) and single-blind in year \preperiod ($\singleblinddisparity{\preperiod}=\disparity{\preperiod}$).

\subsection{Estimation and Inference}
\label{sec:estimation}

Having defined the bias in the review text in terms of observable disparities in Eq.~\eqref{eq:identification}, we now focus on estimating this bias from the available peer review data in years \preperiod and \postperiod. Formalizing the disparities \disparity{\preperiod} and \disparity{\postperiod} in the \textit{text} in a manner that is both substantively meaningful and that permits estimation and inference is non-trivial. In this section, we formalize the disparities in the text and propose a nonparametric procedure to estimate them from the peer reviews in years \preperiod and \postperiod.

Intuitively, the disparity \disparity{t} in the review text in year $t$ is a measure of how the text of the reviews written for \westerngroup differ from those written for \nonwesterngroup. A simple approach to quantify the disparity is to select a ``feature'' of the review text (such as its ``politeness''), annotate the review text based on this feature (either manually or via natural language processing methods) and then compare the value of this feature in the reviews for papers in each of the two subgroups \westerngroup and \nonwesterngroup. However, the disparities and bias quantified in this manner are sensitive to feature selection and annotation.

In contrast, we propose measuring the \disparity{t} nonparametrically, without any feature selection and annotation.
We rely on the intuition that if the text of the reviews written for \westerngroup differs systematically from the text of those written for \nonwesterngroup, a binary machine-learning classifier should be able to distinguish between the reviews written for each subgroup using the \textit{review text}. Hence, we could use any measure of the performance (such as the accuracy, precision, or recall) of such a classifier as a measure of the disparity \disparity{t}.

However, disparities in the review text may also be caused by differences in the research topics pursued by each subgroup.
To ``control for'' subgroup differences in research topics, we rely on the following intuition: subgroup differences in research topics should be reflected in the text of their paper abstracts. Hence, we  quantify subgroup differences in research topics by the ability of a binary machine-learning classifier to distinguish between the papers belonging to each subgroup using their \textit{abstract text}.

We now define the disparity in the review text based on the intuition discussed previously. Let $f(\cdot)$ be a binary classifier mapping a paper's review text to its subgroup and $g(\cdot)$ be a binary classifier mapping a paper's abstract text to its subgroup. Let \perf{f;t} and \perf{g;t} be the chosen measures of classification performance of $f(\cdot)$ and $g(\cdot)$ respectively, such as their area under the ROC curve (AUC), accuracy or precision. We measure the disparity in the review text as the ratio of the performances of the two classifiers on the peer review data in year $t$:
\begin{align}
  \disparity{t} = \perf{f;t} / \perf{g;t}.
  \label{eq:text-disparity}
\end{align}
Normalizing \perf{f;t} by \perf{g;t} as in Eq.~\eqref{eq:text-disparity} ``controls for'' subgroup differences in research topics: if \perf{f;t} is high due to subgroup differences in research topics, \perf{g;t} will also be high. While any binary classifier may be used for $f(\cdot)$ and $g(\cdot)$, poor classifiers are more likely to underestimate the bias
(due to equally poor classification performance in both \preperiod and \postperiod).

In Section~\ref{sec:results}, we report results with multinomial Naive Bayes classifiers for $f(\cdot)$ and $g(\cdot)$ and the AUC as our chosen measure of classification performance. We estimate the value of \perf{f;t} and \perf{g;t} using $k$-fold cross-validation. To eliminate any dependence on the choice of cross-validation folds, we repeat the bias estimation procedure many times with the data belonging to each fold
randomized uniformly in each iteration. We use the empirical distribution of bias estimates from these iterations to construct confidence intervals on the estimated bias.

A final issue we address is that \perf{f;t} and \perf{g;t} can differ in \preperiod and \postperiod due to differences in the sample size (number of reviews) or due to differences in the proportion of papers belonging to each subgroup in \preperiod and \postperiod. Hence, when estimating \disparity{\postperiod} we downsample the available peer review data in year \postperiod\footnote{We do this because the number of reviews in year \postperiod is greater than that in year \preperiod. In general, we downsample the data in the year having a greater number of reviews.} %
such that (i) the number of peer reviews or abstracts is equal to that in \preperiod, (ii) the proportion of abstracts or peer reviews written for papers in subgroup \westerngroup is equal to that in \preperiod, and (iii) the proportion of abstracts or peer reviews written for papers in subgroup \nonwesterngroup is equal to that in \preperiod. As with the cross-validation folds, the downsampling is randomized uniformly in each iteration.

\begin{figure}[t]
  % \vspace{-5mm}
  \centering
  \hspace{-20mm}
  \includegraphics[width=0.55\textwidth]{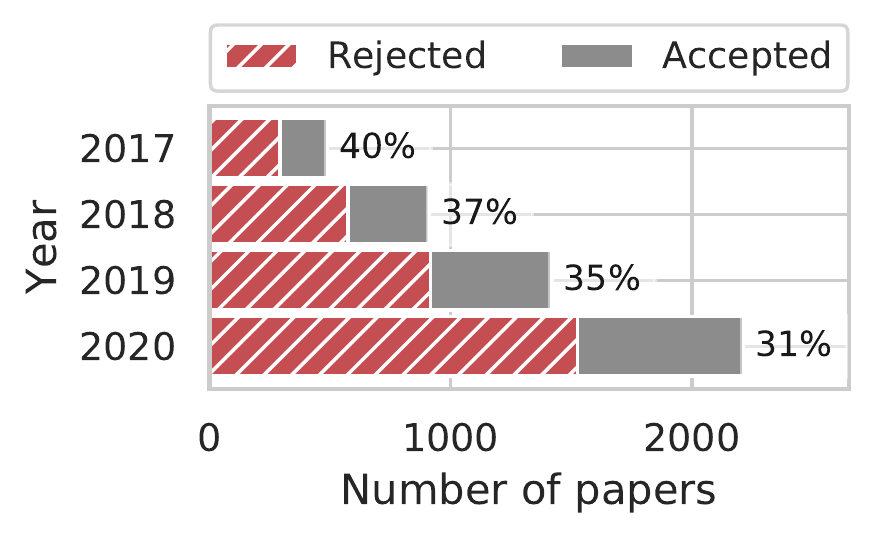}
  \caption{Papers submitted, accepted and rejected from ICLR 2017 through 2020. The proportion of papers accepted in each year is reported to the right of each bar.
  }
  \label{fig:submitted}
\end{figure}
\begin{table}[t]
\centering
  \begin{tabular}{l@{\hskip 4em}c@{\hskip 4em}c}
    \toprule
    & \multicolumn{2}{c}{\textbf{Subgroup Definition}}\\
    \cmidrule(lr){2-3}
    \textbf{Year} & Affiliation-based & Gender-based \\
    \midrule
    2017 & 24.1\% & 25.5\% \\
    2018 & 24.1\% & 28.5\% \\
    2019 & 26.0\% & --- \\
    2020 & 30.5\% & --- \\
    \bottomrule
  \end{tabular}
\caption{Proportion of submitted papers belonging to subgroup \nonwesterngroup in each year for different subgroup definitions.}
\label{tab:groups}
\end{table}

\section{Data}
\label{sec:data}
We assemble a dataset of 16,880 peer reviews from the OpenReview platform for all the 5,638 papers submitted to the International Conference on Learning Representations (ICLR) from 2017 to 2020. Each paper receives 3 peer reviews on average (with a standard deviation of 0.3).
Each peer review contains textual comments and a numerical rating, from 1 to 10 in ICLR 2017--2019 and in \{1, 3, 6, 8\} in ICLR 2020. Fig.~\ref{fig:submitted} reports the number of papers submitted and the proportion of papers accepted in each year.

We investigate the existence of biases in the peer review text with respect to two types of author attributes: (i) the country of their affiliation, and (ii) their perceived gender. Affiliation and gender biases have been a recurring theme in prior work on improving fairness in peer review \citep{blank1991effects,ross2006effect,tomkins2017reviewer,salimi2020causal}. We focus on testing for affiliation and gender bias in the review text to complement prior findings, though biases with respect to other attributes may also exist.

ICLR was co-founded by researchers affiliated with institutions in the USA and Canada. In addition, the general, program and area chairs of ICLR were almost exclusively affiliated with institutions in the USA and Canada since its inception in 2013. Motivated by the possibility of ``in-group bias'' \citep{taylor1981self}, we test for reviewer biases caused by visible author identities in favor of (or against) papers having at least one author with an affiliation in the USA or Canada. We partition the submitted papers into two subgroups, \westerngroup and \nonwesterngroup, based on the countries of the affiliations of their authors. We allocate all papers having at least one author affiliated to a university, organization, or company in the USA or Canada to \westerngroup, and all other papers to \nonwesterngroup. Author affiliations are extracted from the submitted paper PDFs in ICLR 2017, and from the authors' registered emails on the OpenReview platform in ICLR 2018, 2019 and 2020. The goal of this affiliation-based partitioning is to quantify the extent to which reviewers are biased by visible author affiliations.

In addition to affiliation-based subgroups, we consider subgroups based on the authors' gender \textit{as perceived by the reviewer} (and not self-reported).
Since we do not observe how reviewers infer gender from authors' names, we approximate the perceived gender of each author using the following protocol. We first use historical self-reported gender records from the U.S. Social Security Administration to compute the probability of an author's first name being reported as male.\footnote{We do this using the \texttt{gender} package available at https://github.com/ropensci/gender.} If this probability is greater than 90\%, we annotate the author's perceived gender as \textit{male}. If this probability is less than 10\%, we annotate the author's perceived gender as \textit{non-male}. If this probability is between 10\% and 90\%, an external human annotator manually infers
the gender of the author (male or non-male) using visible information on their homepage and Google Scholar profile (found with a Google search). We expect our annotation protocol to approximate the authors' gender as perceived by reviewers\footnote{We designed our annotation protocol to approximate how reviewers perceive the gender of an author from their name without knowledge of the author's self-reported gender. As such, our annotations may contradict authors' gender and cause unintended psychological harm. To prevent this, we do not plan to make our gender annotations public. However, we have described our annotation protocol in sufficient detail for replication.}.

We allocate all papers having at least one author perceived to be non-male to \nonwesterngroup, and those with all authors perceived to be male to \westerngroup. The goal of this gender-based partitioning is to quantify the extent to which reviewers are biased in favor of (or against) papers having at least one author perceived to be non-male. Table~\ref{tab:groups} reports the proportion of papers submitted to ICLR in each year that belong to subgroup \nonwesterngroup for affiliation-based and gender-based subgroup definitions. Since our external human annotations of gender only span ICLR 2017 and 2018, our analyses of bias with gender-based subgroups excludes data from ICLR 2019 and 2020.

A key policy change during this period is ICLR's switch to double-blind reviewing from 2018 onwards. We exploit this policy change to estimate the bias while eliminating the impact of a large class of unobserved confounders, as discussed in Section~\ref{sec:proposed}. We also exploit this policy change in Section~\ref{sec:results}
to construct a ``ground truth'' measure of bias in the peer review ratings using the difference-in-differences methodology. We then evaluate whether the biases estimated by our proposed framework are consistent with the presence and absence of the ``ground truth'' bias in each year.

Since ICLR permits non-anonymized submissions to arXiv and other preprint servers while the paper is under review, it is likely that some author identities were visible even during the double-blind reviewing process in ICLR 2018, 2019 and 2020. Hence, we expect our bias estimates to be conservative (attenuated towards zero).

\section{Evaluation}
\label{sec:results}

We now evaluate the ability of our proposed framework to detect biases in the \emph{text} of peer reviews. Evaluating the validity of causal estimates is challenging in general due to the lack of ``ground truth'' causal effects to compare with, which can only be obtained using randomized control trials that are often expensive and time-consuming. Hence, as in prior work \citep{field2020unsupervised},
evaluation is typically carried out using secondary tasks or semi-synthetic datasets such as the IBM Causal Inference Benchmark \citep{shimoni2018benchmarking}.

Instead of relying on secondary tasks or semi-synthetic datasets that may not represent peer reviewer behavior in the real world, we construct ``ground truth'' bias estimates based on the ratings provided by peer reviewers. Specifically, we apply the difference-in-differences methodology to quantify the presence or absence of biases in the review ratings using peer reviews from each consecutive pair of years between ICLR 2017 and 2020. We then apply our proposed framework to estimate biases in the review text.

We evaluate whether (i) the estimated bias in the review text is statistically significant when the estimated bias in the review ratings is statistically significant, and (ii) the estimated bias in the review text is statistically insignificant when the estimated bias in the review ratings is statistically insignificant. The underlying intuition is that the rating of a review must also be reflected in its text (with language expressing praise or criticism, for example). Hence, an accurate textual bias estimation framework must be able to detect biases (when present) using the review text without having access to the review ratings.

We begin in Section~\ref{sec:groundtruth} with a detailed discussion on estimating the ``ground-truth'' affiliation bias using the difference-in-differences methodology and review ratings. We then estimate and evaluate the affiliation bias in the review text in Section~\ref{sec:bias-in-text}. In Section~\ref{sec:genderbias}, we estimate and evaluate the bias due to perceived gender. In Appendix~\ref{sec:validity}, we further discuss the validity of the identification assumptions used in our evaluation. In Appendix~\ref{sec:alternativedisparity}, as a comparative baseline, we evaluate an alternate measure of disparity in the text proposed by~\citep{gentzkow2019measuring} and empirically show that our proposed measure of disparity has greater statistical power.

\begin{figure}[t]
  % \vspace{-1mm}
  \centering
  \begin{subfigure}[b]{0.495\textwidth}
    \centering
    \includegraphics[width=\textwidth]{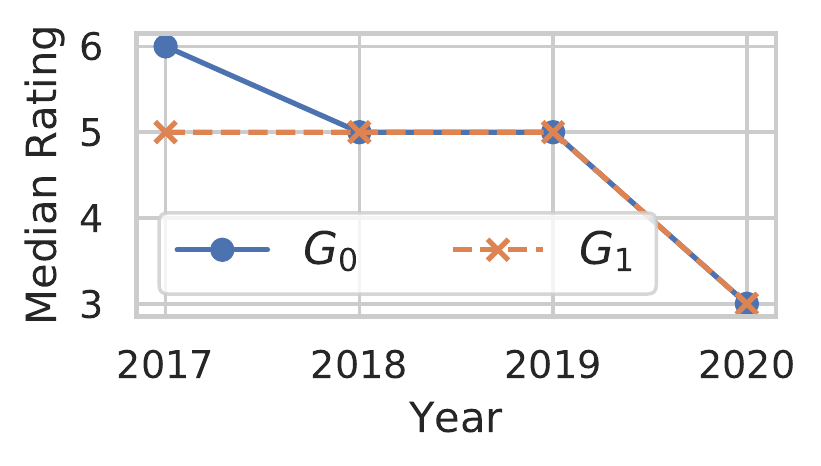}
    \caption{Median review rating for each subgroup}
    \label{fig:medianratings}
  \end{subfigure}
  \hfill
  \begin{subfigure}[b]{0.495\textwidth}
    \centering
    \includegraphics[width=\textwidth]{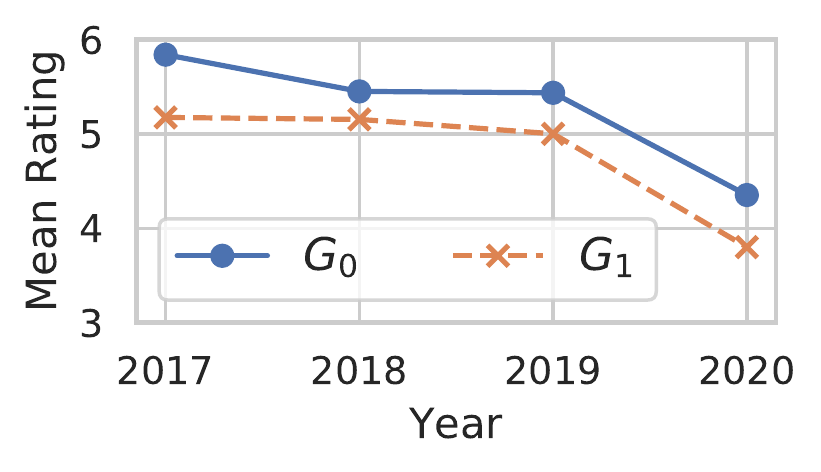}
    \caption{Mean review rating for each subgroup}
    \label{fig:meanratings}
  \end{subfigure}
  \caption{\textbf{Ratings disparities in ICLR 2017 through 2020.} The ratings disparity is quantified by the difference in (a) median, and (b) mean ratings for papers in \westerngroup and \nonwesterngroup.
  }
  \label{fig:modelfree}
\end{figure}
\subsection{Constructing ``ground truth'' affiliation biases from peer review ratings}
\label{sec:groundtruth}

We first provide descriptive evidence in Fig.~\ref{fig:modelfree}
of reviewer bias in favor of papers having at least one author with an affiliation in the USA or Canada (subgroup \westerngroup) in ICLR 2017, when reviewing was single-blind. Fig.~\ref{fig:medianratings} shows that the median review rating for papers in \westerngroup was 1 unit higher than that for papers in \nonwesterngroup in ICLR 2017. This median ratings disparity disappears after the switch to double-blind reviewing in ICLR 2018. Similarly, Fig.~\ref{fig:meanratings} shows that the mean review rating for papers in \westerngroup was 0.665 units higher than that for papers in \nonwesterngroup in ICLR 2017, and 0.297 units higher in ICLR 2018. Thus, switching to double-blind reviewing coincided with a reduction in mean ratings disparity of 0.369 units.

We now estimate the ratings bias using the \textit{difference-in-differences} causal inference methodology \citep{card1994minimum,angrist2008mostly}. The difference-in-differences methodology can be viewed as an analogue of our proposed framework to quantify biases in numerical quantities, instead of biases in unstructured text. This framework has been previously used to quantify gender biases in peer review ratings and hiring decisions \citep{blank91effects,goldin2000orchestrating}, among several other settings.

The difference-in-differences methodology, like our proposed framework,
requires peer reviews in two years \preperiod (with single-blind reviewing) and \postperiod (with double-blind reviewing).
Let \rating be the rating that reviewer $i$ gave to paper $j$, let  $\paperyear \in \{\preperiod, \postperiod\}$ be the year in which paper $j$ was submitted, and let $\papergroup \in \{\westerngroup, \nonwesterngroup\}$ be subgroup that paper $j$ belongs to. The difference-in-differences methodology defines the ratings disparity \ratingsdisparity{t} in each year $t \in \{\preperiod, \postperiod\}$ as:
\begin{align}
  \ratingsdisparity{t} &= \mathbb{E}[\rating | \papergroup = \westerngroup, \paperyear = t]\nonumber\\
                       &-\mathbb{E}[\rating | \papergroup = \nonwesterngroup, \paperyear = t]
  \label{eq:ratingsdisparity}
\end{align}
where the expectations are over all papers $j$ and their respective reviewers $i$. The ratings bias \ratingsbias is defined as the difference in the ratings disparities between the year with single-blind reviewing and the year with double-blind reviewing:
\begin{align}
  \gamma = \ratingsdisparity{\postperiod} - \ratingsdisparity{\preperiod}
  \label{eq:ratingsbias}
\end{align}
Interpreting $\gamma$ as a ratings \textit{bias} -- the difference in mean subgroup ratings \textit{caused} by visible author identities -- requires the \textit{parallel trends} identification assumption. This assumption states that, had the conference never switched to double-blind reviewing, the change in expected rating for subgroup \nonwesterngroup from  \preperiod to \postperiod would have been equal that for subgroup \westerngroup from \preperiod to \postperiod. It is a special case of Assumption~\ref{assumption:parallel-trends} when the disparity in each year is defined as a difference in mean subgroup ratings, as in Eq.~\eqref{eq:ratingsdisparity}. The parallel trends assumption is less restrictive than the ``no unobserved confounders'' assumption.
We discuss the validity of this assumption in our setting in Appendix~\ref{sec:validity}.

The ratings bias $\gamma$ in Eq.~\eqref{eq:ratingsbias} is typically estimated using a ``two-way fixed-effects'' regression \citep{imai2020use} on peer reviews from the years \preperiod and \postperiod:
\begin{align}
  \label{eq:ratingregression}
  \rating  = \rho &+ \alpha \mathbb{I}[\paperyear = \postperiod] +
                       \beta \mathbb{I}[\papergroup = \westerngroup]\\\nonumber
           &+ \gamma \mathbb{I}[\paperyear = \postperiod] \times \mathbb{I}[\papergroup = \westerngroup] +
                          \epsilon_{ij}
\end{align}
where the coefficients $\rho$, $\alpha$, $\beta$ and $\gamma$ are estimated using ordinary least squares (OLS). The error term $\epsilon_{ij}$ is assumed to be Gaussian with zero-mean, which enables deriving asymptotic confidence intervals and p-values for the estimates.

We estimate the ratings bias using the two-way fixed-effects regression model in Eq.~\eqref{eq:ratingregression} on peer reviews from ICLR 2017 (\preperiod) and 2018 (\postperiod). Recall that Fig.~\ref{fig:meanratings}
reports a change in the mean ratings disparity from $\ratingsdisparity{2017}=0.665$ to $\ratingsdisparity{2018}=0.297$, for a total of $\ratingsdisparity{2018}-\ratingsdisparity{2017}=-0.369$ units after switching to double-blind reviewing.
The estimated bias in the first row of Table~\ref{tab:results} (left)
mirrors this. In addition, the confidence intervals and p-value indicate that the estimated bias is statistically significant ($p=0.024$).

Recall that reviewing in ICLR was double-blind in the years 2018, 2019 and 2020. Hence, as ``placebo tests'', we also estimate $\ratingsbias$ using the two-way fixed-effects regression model in Eq.~\eqref{eq:ratingregression}
using peer reviews in the year pair $(\preperiod=2018, \postperiod=2019)$, and the year pair $(\preperiod=2019, \postperiod=2020)$.
The estimates are reported in the second and third rows of Table~\ref{tab:results} (left).
The ratings bias estimated using either of these year pairs is statistically insignificant. This is consistent with the fact that reviewing was double-blind during both years in the pair, and lends support to the validity of the parallel trends assumption.

Table~\ref{tab:results} (left) thus comprises the ``ground truth'' presence and absence of bias for each year pair, which we expect our proposed framework to uncover from the review text without having access to the review ratings.

\begin{table*}[t]
\centering
\small
\begin{tabular}{l@{\hspace{5mm}}ccc @{\hspace{2.5mm}}c@{\hspace{0mm}} ccc}
\toprule
& \multicolumn{3}{c}{\textbf{``Ground truth'' bias in review ratings}}
&
& \multicolumn{3}{c}{\textbf{Estimated bias in the review text}}\\
\cmidrule(lr){2-4}\cmidrule(lr){6-8}
\textrm{Years $\preperiod, \postperiod$} &
\textrm{Bias} &
%\textbf{Std. Error} &
\textrm{p-value} &
\textrm{95\% CI} &
&
\textrm{Bias} &
%\textbf{Std. Error} &
\textrm{p-value} &
\textrm{95\% CI}\\
\midrule
2017, 2018
  & \textbf{-0.369} (0.164) & 0.024 & (-0.690, -0.047)
  &
  & \textbf{-0.166} (0.055) & 0.002 & (-0.270, -0.063)\\[0.8mm]

\textit{Placebo Tests} &&& & &&&\\

\qquad 2018, 2019
  &  \phantom{-}0.138 (0.112) & 0.219 & (-0.082, \phantom{-}0.358)
  &
  &  -0.070  (0.068) & 0.308 & (-0.195, \phantom{-}0.072)\\

\qquad 2019, 2020
  & \phantom{-}0.118 (0.099) & 0.236 & (-0.077, \phantom{-}0.313)
  &
  & \phantom{-}0.012  (0.043) & 0.781 & (-0.082, \phantom{-}0.083)\\

\bottomrule
\end{tabular}
\caption{\textbf{``Ground truth'' and estimated bias with respect to affiliation.} ``Ground truth'' difference-in-difference estimates of the bias in the review ratings (left) and bias in the review text estimated by our proposed framework (right). Standard errors reported in brackets. Estimates in each row are computed using ICLR peer reviews in consecutive years $\preperiod$ and $\postperiod$. Estimates in bold are statistically significant at the 5\% level.
}
\label{tab:results}
% \vspace{-1mm}
\end{table*}

\subsection{Estimating and evaluating affiliation bias in the review text}
\label{sec:bias-in-text}

We now estimate the bias in the review text using equations \eqref{eq:bias} and \eqref{eq:text-disparity}. We use multinomial Naive Bayes classifiers with add-one smoothing for $f(\cdot)$ and $g(\cdot)$ on frequencies of unigrams and bigrams in the review and abstract text respectively. We use the area under the ROC curve (AUC) for both \perf{f;t} and \perf{g;t}, estimated using 10-fold cross-validation. We downsample the reviews and abstracts in year \postperiod to equalize the sample sizes and subgroup proportions in \preperiod and \postperiod, as described in Section~\ref{sec:proposed}. We repeat the bias estimation procedure 1,000 times with downsampling and the cross-validation folds randomized uniformly in each iteration. We use the empirical distribution of bias estimates from these iterations to construct confidence intervals on the estimated bias. We compute the p-value from the confidence intervals using the analytical method proposed by \citet{altman2011obtain}.

The estimated biases in the review text for the year pair $(\preperiod=2017, \postperiod=2018)$ and the placebo year pairs $(\preperiod=2018, \postperiod=2019)$ and $(\preperiod=2019, \postperiod=2020)$ are reported in Table~\ref{tab:results} (right). The first row of Table~\ref{tab:results} (right) reports a statistically significant ($p = 0.002$) estimated bias corresponding to a \textit{reduction} of 0.166 units (and hence, a negative estimate) in the classification performance ratio (see Eq.~\ref{eq:text-disparity}) from ICLR 2017 to 2018. The second and third rows of Table~\ref{tab:results} (right) report statistically insignificant bias estimates using peer reviews in the double-blind year pairs $(\preperiod=2018, \postperiod=2019)$ and $(\preperiod=2019, \postperiod=2020)$. The biases in the review text in each year pair estimated using our proposed framework are consistent with the presence and absence of ``ground truth'' ratings bias in each year pair reported in Table~\ref{tab:results} (left). This validates the effectiveness of our proposed framework.

\subsection{Estimating and evaluating bias with respect to the authors' perceived gender}
\label{sec:genderbias}

Different types of biases may be expressed in the text against population subgroups defined in different ways (such as by affiliation, race or gender). Our proposed framework does not rely on linguistic feature-engineering targeted at any specific type of bias.
Hence, we evaluate
the ability of our proposed framework to test
for biases with subgroups defined based on the authors' gender \textit{as perceived by the reviewer} (and not their self-reported gender). We detailed our gender-based subgroup definition and our gender annotation protocol earlier in Section~\ref{sec:data}.

Since our manual gender annotations only span ICLR 2017 and 2018, we report the estimated bias in the review ratings and text using ICLR 2017 and 2018 in Table~\ref{tab:bias-gender}. The ``ground truth'' bias in the review ratings (estimated using the difference-in-differences methodology as in Section~\ref{sec:groundtruth}) is statistically insignificant. The bias in the review text estimated using our proposed framework is also statistically insignificant, and hence, consistent with the ``ground truth''.

In summary, given our data and choice of gender-based subgroups, we cannot reject the null hypotheses of there being no bias in the review ratings and text against papers with at least one author perceived to be non-male. It is, however, important to note that failing to reject the null hypothesis does not confirm the absence of gender bias.

\begin{table}[t]
\centering
%\small
\begin{tabular}{lccc}
\toprule
\textbf{Source} &
\textbf{Bias} &
%\textbf{Std. Error} &
\textbf{p-value} &
\textbf{95\% CI}\\
\midrule
Ratings & -0.073 (0.160) & 0.647 & (-0.386, 0.240)\\
Text    & -0.468 (0.335) & 0.163 & (-0.862, 0.198)\\
\bottomrule
\end{tabular}
\caption{\textbf{``Ground truth'' and estimated bias with respect to perceived gender.} Estimated bias in the review text and review ratings with respect to the authors' perceived gender using ICLR peer reviews in the years 2017 and 2018. Standard errors reported in brackets.}
\label{tab:bias-gender}
%\vspace{-4mm} % INCREASE TO MAKE SPACE FOR ACKs
\end{table}

\clearpage

\section{Conclusion and Discussion}
\label{sec:conclusion}
Our work addresses an important yet relatively overlooked medium through which biases can harm society --- that of text-based communication. We propose a framework to nonparametrically estimate  biases expressed in text, which is robust to a larger class of unobserved confounders than prior work. We evaluate our approach in the setting of scholarly peer-review, wherein the ``ground truth'' bias can be inferred, and show that our proposed framework detects bias in the peer review text that is consistent with the ``ground truth''.
Our framework can be used by policymakers to formulate more effective bias-mitigation policies that improve the equitability of hiring, promotion and other socioeconomic processes.

More generally, our work extends the difference-in-differences methodology to accommodate unstructured text as the ``outcome''. It operates on text observed in two time periods associated with two population subgroups before and after a (potentially identity-hiding) policy change, such as switching to age-blind recruitment \citep{capowski1994ageism}, blind performance reviews \citep{goldin2000orchestrating} or blind grading \citep{hanna2012discrimination}. Our proposed framework quantifies the \textit{causal} effect of the policy change on the difference in the text associated with each population subgroup. As such, our work also contributes to the nascent literature on causal inference from text \citep{roberts2018adjusting,egami2018make,pryzant2018interpretable,sridhar2019estimating,keith2020text} with ``text as the outcome''.

As a policymaking tool, one should take care in using our proposed framework appropriately. 
Our evaluation study in Section~\ref{sec:results} is a detailed example of how the estimates and confidence intervals from our proposed framework are to be interpreted, both when they are statistically significant and insignificant. We expect that, with this example, users of our proposed framework are motivated to employ similar care when interpreting the bias estimates in their setting.

A correct use of our proposed framework would entail carefully considering the validity of the underlying identification assumption (Assumption~\ref{assumption:parallel-trends}). 
If an unobserved confounder exists that violates Assumption~\ref{assumption:parallel-trends}, our estimates will quantify the change in disparity from \preperiod to \postperiod caused by a combination of hiding author identities \emph{and} the unobserved confounder, which cannot be interpreted as bias. While this assumption is empirically untestable (since it involves unobserved counterfactual quantities), placebo tests can be used to empirically support its validity. However, note that while the failure of a placebo test implies that the identification assumption does not hold, the success of a placebo test does {not} confirm that the identification assumption holds.

In our peer review setting, a potential confounder is an increase in research funding \emph{only} for the institutions comprising \nonwesterngroup from ICLR 2017 to 2018, and not for those comprising \westerngroup. In Fig.~\ref{fig:modelfree}, note that the subgroup disparity in the mean and median review ratings decreases from ICLR 2017 to 2018 (the mean and median review ratings for the two subgroups become more similar in 2018). However, also note that the reduction in disparity is due to a \emph{decrease} in the mean and median ratings for subgroup \westerngroup in 2018. Had research funding for the institutions comprising \nonwesterngroup increased, we would have expected an \emph{increase} in the mean or median ratings for subgroup \nonwesterngroup in 2018. Hence, the temporal trends in ratings in Fig.~\ref{fig:modelfree} contradict the hypothesis of confounding due to an increase in research funding for the institutions comprising \nonwesterngroup. In Appendix~\ref{sec:validity}, we detail this argument further and show how a combination of substantive reasoning, empirical tests and external evidence must be used to assess the validity of the identification assumption.

\section*{Acknowledgments} 
The work of Nihar Shah was supported by an NSF CAREER Award CIF: 1942124.

\setlength{\bibsep}{3pt plus 0.1ex}
\begin{small}
\bibliography{refs}
\end{small}

%\clearpage

\appendix

\section{Validity of the identification assumptions}
\label{sec:validity}

Our evaluation in Section~\ref{sec:results} relies on Assumption~\ref{assumption:parallel-trends} (required by our proposed framework) and the parallel trends assumption (required by the difference-in-differences methodology used to construct the ``ground truth''). Both assumptions are robust to unobserved confounders that (i) do not change from \preperiod to \postperiod, or (ii) change from \preperiod to \postperiod, but affect the peer review text or rating for both subgroups to the same extent. For example, if the peer reviewers were harsher in \postperiod than in \preperiod by the same amount for \westerngroup and \nonwesterngroup, the average peer review ratings and the ``positivity'' of the peer review text would be lower in \postperiod for both \westerngroup and \nonwesterngroup by the same amount. Hence, having harsher reviewers in \postperiod would have no effect on the review text or ratings disparities that comprise the bias in Eqs.~\eqref{eq:bias} and \eqref{eq:ratingsbias}.

If an unobserved confounder exists that violates these assumptions, our estimates will quantify the change in disparity from \preperiod to \postperiod caused by a combination of hiding author identities \emph{and} the unobserved confounder. Neither assumption is empirically testable (since each involves unobserved counterfactual quantities). Placebo tests can empirically support (though not confirm) the validity of these assumptions. Hence, the identification assumptions must be argued for substantively (with external evidence if possible), and care must be taken when interpreting the estimates as biases (the change in disparity \textit{caused} by hiding author identities).

In our setting, a potential confounder is an increase in research funding \emph{only} for the institutions comprising \nonwesterngroup from ICLR 2017 to 2018, and not for those comprising \westerngroup. Recall that papers in \nonwesterngroup were rated lower on average than those in \westerngroup in ICLR 2017.
A funding increase of this type could result in an improvement in the average quality of papers in \nonwesterngroup from 2017 to 2018, with no change in the average quality of papers in \westerngroup. This would, in turn, cause a reduction in the subgroup disparity from ICLR 2017 to 2018. Hence, if such a confounder was present, it would be impossible to attribute the change in disparity from ICLR 2017 to 2018 entirely to the switch to double-blind reviewing.

We argue against the existence of such a confounder. In Fig.~\ref{fig:modelfree}, we note that the disparity in the mean and median review ratings for each subgroup decreases from ICLR 2017 to 2018. However, we also note that this reduction is due to a \emph{decrease} in the mean and median ratings for subgroup \westerngroup. Had the aforementioned confounder been present, we would expect the disparity to reduce due to an \emph{increase} in the mean or median ratings for subgroup \nonwesterngroup. Hence, this contradicts the claim that an increase in research funding only for the institutions comprising \nonwesterngroup confounds our bias estimates.

Another potential confounder is a \emph{decrease} in research funding for the institutions comprising \westerngroup, which would lead to a reduction in disparity from ICLR 2017 to 2018 (if the quality of papers in \westerngroup decreased due to lesser funding). However, federal research funding in the USA increased in every year from 1953 to 2018 \citep{nsf2020}, and federal research funding in Canada increased in every year since 2013 \citep{can2020}.
Thus, it is unlikely that a decrease in research funding for the institutions comprising \westerngroup confounds our bias estimates.

\begin{table}[t]
\centering
% \small
\begin{tabular}{l@{\hspace{3mm}}c@{\hspace{3mm}}c@{\hspace{3mm}}c@{\hspace{3mm}}c}
\toprule
\textbf{Years $\preperiod, \postperiod$} &
\textbf{Bias} &
%\textbf{Std. Error} &
\textbf{p-value} &
\textbf{95\% CI}\\
\midrule
2017, 2018        & -0.468 (0.335) & 0.163 & (-0.862, 0.198)\\[0.8mm]
% \textit{Placebo Tests} &&&&\\
2018, 2019 &  \phantom{-}0.549  (0.376) & 0.145 & (-0.025, 0.864)\\
2019, 2020 &        -0.501  (0.384) & 0.193 & (-0.835, 0.091)\\
\bottomrule
\end{tabular}
\caption{\textbf{Affiliation bias estimated using \citep{gentzkow2019measuring}.}
Estimated affiliation biases in the review text using ICLR peer reviews in consecutive years \preperiod and \postperiod. The disparity in the text is estimated using the method proposed by \citet{gentzkow2019measuring}. Standard errors are reported in brackets.
 }
\label{tab:gentzkow}
% \vspace{-1mm}
\end{table}
\section{Comparison with an alternate measure of disparity in the text}
\label{sec:alternativedisparity}
Recall from Eq.~\eqref{eq:text-disparity} that our proposed framework quantifies disparity in the text using the performance of binary classifiers in predicting subgroup membership. We now evaluate an alternate method to measure disparity in the text proposed by \citet{gentzkow2019measuring}. We estimate the bias in the review text using the disparities derived from this method for each pair of consecutive years from ICLR 2017 to 2020, and compare these estimates with the ``ground truth'' biases estimated in Section~\ref{sec:groundtruth}.

\citet{gentzkow2019measuring} propose measuring subgroup disparities in text by estimating a regularized Poisson generative model of word frequencies as a function of subgroup membership. Specifically, the disparity for each word in the review (or abstract) text is estimated using an $L_1$-regularized Poisson regression of the word frequency on the subgroup membership indicator. The magnitude of the regularization penalties are chosen to minimize the Bayesian Information Criterion (BIC) of the estimated model. The overall disparity in the review (or abstract) text is computed by aggregating all word-level disparities. Confidence intervals are constructed by repeated estimation using random subsampling \textit{without} replacement.

We estimate the bias in the review text using equations \eqref{eq:bias} and \eqref{eq:text-disparity}, with  \perf{f;t} and \perf{g;t} replaced by disparities in the review and abstract text (respectively) estimated using the method proposed by \citet{gentzkow2019measuring}. As in Section~\ref{sec:bias-in-text}, we downsample the reviews and abstracts in year \postperiod and repeat the bias estimation procedure 1,000 times to construct confidence intervals. The estimated biases are reported in Table~\ref{tab:gentzkow}.
For any year pair, with disparities in the text estimated using the method proposed by \citet{gentzkow2019measuring}, we cannot reject the null hypothesis of there being no bias in the review text.

Hence, compared to our proposed method to measure disparity in the text, the method proposed by \citet{gentzkow2019measuring} lacks statistical power for the data used in our setting. We attribute this to two factors. First, the frequencies of most words used in the reviews and abstracts are low, which would lead to noisy estimates of word-level disparities, and hence, noisy overall disparity estimates. Second, the Poisson regression imposes strong distributional assumptions on how the text is generated that are unlikely to hold in all settings.

\end{document}